\documentclass[letterpaper, 10 pt, conference]{ieeeconf}
\IEEEoverridecommandlockouts
\usepackage[utf8]{inputenc}
\usepackage{graphicx}
\usepackage{multicol}
\usepackage{footmisc}
\usepackage{amsfonts}

\usepackage{subcaption}
\usepackage{amsmath}
\usepackage{amssymb}
\usepackage[english]{babel}
\usepackage[maxbibnames=99,natbib=true,style=numeric-comp,backend=biber,sorting=nty]{biblatex}
\usepackage{csquotes}
\let\labelindent\relax
\usepackage[inline]{enumitem}
\usepackage{todonotes}
\usepackage[normalem]{ulem}
\usepackage{version}
\usepackage{booktabs}
\usepackage{xcolor}
\usepackage[ruled,vlined,linesnumbered,noend]{algorithm2e}
\usepackage{mathtools}
\usepackage{tcolorbox}
\usepackage[framemethod=tikz]{mdframed}
\usepackage[bookmarks=true]{hyperref}
\let\subparagraph\relax
\usepackage[font=small,skip=0pt]{caption}
\let\proof\relax
\let\endproof\relax
\usepackage{amsthm}
\usepackage[nameinlink,capitalize]{cleveref}

\usepackage{xcolor}

\newtheorem{theorem}{Theorem}
\newtheorem{corollary}{Corollary}
\newtheorem{lemma}{Lemma}

\crefname{lemma}{Lemma}{Lemmas}
\Crefname{lemma}{Lemma}{Lemmas}

\crefformat{footnote}{#2\footnotemark[#1]#3}

\newcounter{note}
\newcounter{assume}
\newcounter{sublemma}
\newcounter{observation}

\crefformat{app}{#2Appendix #1#3}
\crefrangeformat{app}{#3Appendices #1#4 to #5#2#6}
\crefmultiformat{app}{#2Appendices #1#3}{ and #2#1#3}{, #2#1#3}{ and #2#1#3}
\crefrangemultiformat{app}{#3Appendices #1#4 to #5#2#6}{ and #3#1#4 to #5#2#6}{, #3#1#4 to #5#2#6}{ and #3#1#4 to #5#2#6}
\Crefformat{app}{#2Appendix #1#3}
\Crefrangeformat{app}{#3Appendices #1#4 to #5#2#6}
\Crefmultiformat{app}{#2Appendices #1#3}{ and #2#1#3}{, #2#1#3}{ and #2#1#3}
\Crefrangemultiformat{app}{#3Appendices #1#4 to #5#2#6}{ and #3#1#4 to #5#2#6}{, #3#1#4 to #5#2#6}{ and #3#1#4 to #5#2#6}

\crefformat{note}{#2note #1#3}
\crefrangeformat{note}{#3notes #1#4 to #5#2#6}
\crefmultiformat{note}{#2notes #1#3}{ and #2#1#3}{, #2#1#3}{ and #2#1#3}
\crefrangemultiformat{note}{#3notes #1#4 to #5#2#6}{ and #3#1#4 to #5#2#6}{, #3#1#4 to #5#2#6}{ and #3#1#4 to #5#2#6}
\Crefformat{note}{#2Note #1#3}
\Crefrangeformat{note}{#3Notes #1#4 to #5#2#6}
\Crefmultiformat{note}{#2Notes #1#3}{ and #2#1#3}{, #2#1#3}{ and #2#1#3}
\Crefrangemultiformat{note}{#3Notes #1#4 to #5#2#6}{ and #3#1#4 to #5#2#6}{, #3#1#4 to #5#2#6}{ and #3#1#4 to #5#2#6}

\crefformat{assume}{#2assumption #1#3}
\crefrangeformat{assume}{#3assumptions #1#4 to #5#2#6}
\crefmultiformat{assume}{#2assumptions #1#3}{ and #2#1#3}{, #2#1#3}{ and #2#1#3}
\crefrangemultiformat{assume}{#3assumptions #1#4 to #5#2#6}{ and #3#1#4 to #5#2#6}{, #3#1#4 to #5#2#6}{ and #3#1#4 to #5#2#6}
\Crefformat{assume}{#2Assumption #1#3}
\Crefrangeformat{assume}{#3Assumptions #1#4 to #5#2#6}
\Crefmultiformat{assume}{#2Assumptions #1#3}{ and #2#1#3}{, #2#1#3}{ and #2#1#3}
\Crefrangemultiformat{assume}{#3Assumptions #1#4 to #5#2#6}{ and #3#1#4 to #5#2#6}{, #3#1#4 to #5#2#6}{ and #3#1#4 to #5#2#6}

\crefformat{sublemma}{#2sub-lemma #1#3}
\crefrangeformat{sublemma}{#3sub-lemmas #1#4 to #5#2#6}
\crefmultiformat{sublemma}{#2sub-lemmas #1#3}{ and #2#1#3}{, #2#1#3}{ and #2#1#3}
\crefrangemultiformat{sublemma}{#3sub-lemmas #1#4 to #5#2#6}{ and #3#1#4 to #5#2#6}{, #3#1#4 to #5#2#6}{ and #3#1#4 to #5#2#6}
\Crefformat{sublemma}{#2Sub-lemma #1#3}
\Crefrangeformat{sublemma}{#3Sub-lemmas #1#4 to #5#2#6}
\Crefmultiformat{sublemma}{#2Sub-lemmas #1#3}{ and #2#1#3}{, #2#1#3}{ and #2#1#3}
\Crefrangemultiformat{sublemma}{#3Sub-lemmas #1#4 to #5#2#6}{ and #3#1#4 to #5#2#6}{, #3#1#4 to #5#2#6}{ and #3#1#4 to #5#2#6}

\crefformat{observation}{#2observation #1#3}
\crefrangeformat{observation}{#3observations #1#4 to #5#2#6}
\crefmultiformat{observation}{#2observations #1#3}{ and #2#1#3}{, #2#1#3}{ and #2#1#3}
\crefrangemultiformat{observation}{#3observations #1#4 to #5#2#6}{ and #3#1#4 to #5#2#6}{, #3#1#4 to #5#2#6}{ and #3#1#4 to #5#2#6}
\Crefformat{observation}{#2Observation #1#3}
\Crefrangeformat{observation}{#3Observations #1#4 to #5#2#6}
\Crefmultiformat{observation}{#2Observations #1#3}{ and #2#1#3}{, #2#1#3}{ and #2#1#3}
\Crefrangemultiformat{observation}{#3Observations #1#4 to #5#2#6}{ and #3#1#4 to #5#2#6}{, #3#1#4 to #5#2#6}{ and #3#1#4 to #5#2#6}

\tcbset{colback=gray!5,colbacktitle=gray!20,coltitle=black}
\setlist[description]{leftmargin=\parindent}

\hypersetup{
	colorlinks=true,
	linkcolor={red!50!black},
	citecolor={green!80!black},
	urlcolor={blue!80!black},
	pdftitle={Ensuring Progress for Multiple Mobile Robots via Space Partitioning, Motion Rules, and Adaptively Centralized Conflict Resolution}
	pdfauthor={Claire Liang, Wil Thomason, E. Andy Ricci, and Soham Sankaran},
	pdfsubject=Multi-agent Coordination,
	pdfkeywords=Multi-robot Coordination;Multi-agent Coordination;Robotics;Planning
}

\newbool{comments}
\newcommand{\wtnote}[2][noinline]{\ifbool{comments}{\todo[author=WT,color=green!40,#1]{#2}}{}}
\newcommand{\clnote}[2][noinline]{\ifbool{comments}{\todo[author=CL,color=orange!40,#1]{#2}}{}}
\newcommand{\ernote}[2][noinline]{\ifbool{comments}{\todo[author=ER,color=cyan!40,#1]{#2}}{}}
\newcommand{\ssnote}[2][noinline]{\ifbool{comments}{\todo[author=SS,color=yellow!40,#1]{#2}}{}}

\newbibmacro{string+url}[1]{%
	\iffieldundef{url}{#1}{\href{\thefield{url}}{#1}}}
\DeclareFieldFormat{title}{\usebibmacro{string+url}{\mkbibemph{#1}}}
\DeclareFieldFormat[article]{title}{\usebibmacro{string+url}{\mkbibquote{#1}}}
\DeclareFieldFormat[inproceedings]{title}{\usebibmacro{string+url}{\mkbibquote{#1}}}
\DeclareFieldFormat[book]{title}{\usebibmacro{string+url}{\mkbibquote{#1}}}
\DeclareFieldFormat[incollection]{title}{\usebibmacro{string+url}{\mkbibquote{#1}}}
\newcommand{\VelBound}{\ensuremath{\mathbf{V}}}

\newcommand{\environment}{\ensuremath{\mathcal{E}}}

\newcommand{\freespace}{\ensuremath{\mathcal{F}}}

\newcommand{\ball}{\ensuremath{\mathcal{B}}}

\newcommand{\idx}{\ensuremath{i}}
\newcommand{\robot}{\ensuremath{r_\idx}}
\newcommand{\robotState}{\ensuremath{\robot(t)}}
\newcommand{\velState}{\ensuremath{\vec{v}_i(t)}}
\newcommand{\robots}{\ensuremath{\mathcal{R}}}
\newcommand{\openSpace}{\ensuremath{\Omega}}
\newcommand{\flowSpace}{\ensuremath{\Phi}}
\newcommand{\passageSpace}{\ensuremath{\Psi}}
\newcommand{\openRegion}{\ensuremath{O}}
\newcommand{\flowRegion}{\ensuremath{F}}
\newcommand{\passageRegion}{\ensuremath{P}}
\newcommand{\spot}{\ensuremath{B_{\radius}}}
\newcommand{\timestamp}{\ensuremath{\tau}}
\newcommand{\prioritybreaker}{\ensuremath{\ell}}
\newcommand{\spaceRequest}{\ensuremath{SR}}

\newcommand{\priority}{\ensuremath{L}}
\newcommand{\radius}{\ensuremath{\rho}}

\newcommand{\god}{\ensuremath{\chi}}

\newcounter{ruledefs}

\crefname{ruledefs}{Rule}{Rules}

\newenvironment{sketch}{%
  \proof}{\endproof}

\title{Ensuring Progress for Multiple Mobile Robots via Space Partitioning, Motion Rules, and Adaptively Centralized Conflict Resolution}
\author{Claire Liang$^{1,*}$ \and Wil Thomason$^{2,*}$ \and E. Andy Ricci$^1$ \and Soham Sankaran$^{1,3}$%
\thanks{$^1$: Department of Computer Science,
Cornell University,
Ithaca, NY, USA 
\texttt{\footnotesize \{cliang,ericci\}@cs.cornell.edu}}%
\thanks{$^2$: Department of Computer Science,
Rice University,
Houston, TX, USA.
Work done while at $^1$.
\texttt{\footnotesize wbthomason@rice.edu}}%
\thanks{$^3$: Pashi Corp., \texttt{\footnotesize soham@soh.am}}%
\thanks{$^*$ These authors contributed equally to this work.}}

\begin{document}
\maketitle

\begin{abstract}
  In environments where multiple robots must coordinate in a shared space, decentralized approaches allow for decoupled planning at the cost of global guarantees, while centralized approaches make the opposite trade-off. These solutions make a range of assumptions --- commonly, that all the robots share the same planning strategies. 
  In this work, we present a framework that ensures progress for all robots without assumptions on any robot's planning strategy by
  \begin{enumerate*}[label=(\arabic*)]
    \item generating a partition of the environment into ``flow'', ``open'', and ``passage'' regions and
    \item imposing a set of rules for robot motion in these regions
\end{enumerate*}.
  These rules for robot motion prevent deadlock through an adaptively centralized protocol for resolving spatial conflicts between robots.
  Our proposed framework ensures progress for all robots without a grid-like discretization of the environment or strong requirements on robot communication, coordination, or cooperation.
  Each robot can freely choose how to plan and coordinate for itself, without being vulnerable to other robots or groups of robots blocking them from their goals, as long as they follow the rules when necessary.
  We describe our space partition and motion rules, prove that the motion rules suffice to guarantee progress in partitioned environments, and demonstrate several cases in simulated polygonal environments. This work strikes a balance between each robot's planning independence and a guarantee that each robot can always reach any goal in finite time.
\end{abstract}

\section{Introduction}\label{sec:introduction}
Some of the most successful real-world applications of robotics, including deployments in warehouses and industrial settings, require coordinating multiple mobile robots in a shared space to avoid conflicts. 
Most approaches to multi-robot coordination in these settings assume that all robots share the same planning strategies.
While this assumption may hold in real-world spaces today, in the future we expect that mobile robots controlled by different entities and equipped with different planning strategies will have to coexist safely in spaces such as shared warehouses, docks, and offices such that each robot is able to access shared resources and make progress. 

In this paper, we introduce the concept of a \emph{motion pact}, which lays out a set of rules for motion in a shared space independent of any particular planning strategy and provides certain guarantees if all robots in the space follow those rules.
We propose one such motion pact which guarantees progress for all robots while imposing relatively lightweight restrictions on each robot's space of planning choices, ensuring that each robot can largely plan, coordinate, and move independently from other robots in a decentralized fashion as long as it obeys the rules when necessary.
The motion pact achieves this by
\begin{enumerate*}[label=(\arabic*)]
\item dividing up the shared environment into regions (\cref{sec:method.spacetypes}), and
\item installing a set of motion rules (\cref{sec:method.motionrules}) that consist of guided motion in certain regions and an \emph{online adaptive centralization mechanism} for resolving spatial conflicts (\cref{sec:method.spacerequests})
\end{enumerate*}
We prove that progress is guaranteed for all robots if they all honor the motion rules (\cref{sec:analysis}) --- in particular, that at any instant, any robot is able to access any location in finite time --- and demonstrate this practically via a proof-of-concept simulation implementation (\cref{sec:evaluation}). 

\section{Related Work}\label{sec:relatedwork}
Multi-robot path planning and coordination has been studied extensively~\cite{honig_persistent_robust_2019,bennewitz_finding_optimizing_2002,buckley_fast_motion_1989,erdmann_multiple_moving_1986,jager_decentralized_collision_2001,kato_coordinating_mobile_1992,fainekos_temporal_logic_2009,yu_multi-agent_path_2013,yan_survey_analysis_2013,sharon_conflict-based_search_2012,barer_suboptimal_variants_2014,krishna_reactive_navigation_2005,velagapudi_decentralized_prioritized_2010,luna_efficient_complete_2011,svestka_coordinated_path_1998,saha_multi-robot_motion_2006,vandenberg_centralized_path_2009,vandenberg_prioritized_motion_2005,clark_motion_planning_2003,vandenberg_reciprocal_velocity_2008,vedder2019x,wagner_mstar:_complete_2011}.
Much of this work can be characterized by its degree of \emph{centralization}: fully centralized methods~\cite{arai_motion_planning_1992,sharon_conflict-based_search_2012,barer_suboptimal_variants_2014,vandenberg_centralized_path_2009} tend to offer guarantees around \emph{progress} (i.e.\ ensuring that all robots reach their goals) and \emph{optimality} with regards to path length, but are typically computationally expensive and limited in their scalability.
At the other end of this spectrum, fully decentralized methods~\cite{jager_decentralized_collision_2001,velagapudi_decentralized_prioritized_2010,pallottino_decentralized_cooperative_2007,vandenberg_reciprocal_velocity_2008} most often take the opposite trade-off: high scalability, but weaker guarantees of progress and optimal performance.
Many approaches, including ours, lie between these extremes; we call these \emph{semi-centralized}~\cite{oh_centralized_decoupled_2011,clark_motion_planning_2003,wagner_mstar:_complete_2011, simmons2000coordination}.

Our approach permits robots to remain fully decentralized and decoupled until a conflict arises during execution, at which time a subset of the robots must coordinate.
It combines and builds on work in semi-centralized planning and space partitioning to offer guaranteed progress without making strong assumptions on robot planning methods.
Our space partition prevents deadlock in narrow passageways, ensuring that every point in the workspace will always be safely reachable in finite time.
We resolve conflicts in tightly-packed space with a protocol that builds on priority-based planning~\cite{velagapudi_decentralized_prioritized_2010} and local conflict resolution~\cite{li_motion_planning_2003, clark_motion_planning_2003, krishna_reactive_navigation_2005, silver2005cooperative}.
We also guarantee progress and are complete; we do not provide any guarantee on path length optimality.

\Citet{vandenberg_centralized_path_2009} is similar to our approach in that it performs maximally decoupled centralized planning for a set of robots by computing optimal partitions of robots to centrally coordinate.
Their algorithm runs offline and creates sequentially executed plans among the partitioned subsets of robots.
In contrast, our approach is designed to work online: we do not explicitly partition robots, and instead induce an implicit partition on-the-fly, selecting robots as-needed to coordinate to resolve a given conflict.

Finally, most existing automated warehouse systems use fully centralized methods with all robots operating under a single unified planner~\cite{wurman_coordinating_hundreds_2008,dandrea_guest_editorial:_2012}.
Our proposed approach has particular benefits for \emph{shared warehouses} and other spaces in which multiple distinct entities (e.g.\ companies) have distinct, independently-operated teams of robots operating in a single shared space.
Unlike existing multi-robot coordination methods, our proposed approach can accommodate these entities using their own, separate planners for their robots, requiring only that the robots coordinate when conflicts arise, while ensuring that each robot can access every shared resource in finite time infinitely many times.
% \wtnote[inline]{Gotta have more to justify this claim in comparison to other approaches. Also still need to mention~\cite{yan_survey_analysis_2013,yu_multi-agent_path_2013,peasgood_complete_scalable_2008,li_motion_planning_2003,li_motion_planning_2005,alami_multi-robot_cooperation_1995,krishna_reactive_collision_2004,vandenberg_centralized_path_2009,fainekos_temporal_logic_2009,kloetzer_fully_automated_2008}.}

%\wtnote[inline]{These section titles aren't meant to be final, just for organizing things to cite.}
%\subsection{Warehouse Robotics}
%\begin{enumerate}
% \item Multi-robot systems are useful in warehouses, but existing warehouse systems use a unified, centralized planner and cannot handle agents operating in the same space not controlled by the centralized planner~\cite{reuters_fight_amazon_2016, staff_what_robots_2019, wurman_coordinating_hundreds_2008, dandrea_guest_editorial:_2012, honig_persistent_robust_2019}.
%\end{enumerate}

\section{Problem Setup}
Let \robots{} be a finite set of omnidirectional robots in a two-dimensional polygonal environment \environment{} consisting of free space \freespace{} and polygonal obstacles $O$.
Assume each robot $\robot \in \robots$ is modeled as a disc with the same fixed radius $\radius$, and that each $\robot \in \robots$ 
\begin{enumerate*}[label=(\arabic*)]
    \item\label{assumption:speedlimit} has a maximum speed bounded by $\VelBound \in \mathbb{R}^+$ and 
    \item\label{assumption:safety} uses a local safety controller to never overlap with any obstacle in $O$ or other robot $r_j \in \robots$, $i \ne j$
\end{enumerate*}.
Finally, let each robot be independently controlled, with its own navigation strategy and goals, and initialize the $\robot \in \robots$ to unique, non-overlapping positions in \environment{} at timestep $t = 0$.

We wish to guarantee that all $\robot \in \robots$ can safely navigate to their goals in finite time.
%\input{sections/assumptions.tex}

%\section{Method}\label{sec:method}

\section{Space Partitioning: Regions}\label{sec:method.spacetypes}
We partition the free space of an environment into three classes of space: \emph{flow space}, \emph{open space}, and \emph{passage space} according to the algorithm described in~\cref{appendix:B}.
These classes are composed of, respectively, \emph{flow regions}, \emph{open regions}, and \emph{passage regions}.
\label{definition:spot}%
In the following, a \emph{``spot''} refers to a \(\ball(\radius) \subseteq \freespace\), an exactly robot-shaped disc in the free space of \environment{}.
As disc-modeled robots in arbitrary environments can become stuck when the space is too densely occupied, we impose a global density cap:

\refstepcounter{assume}
\begin{tcolorbox}[title=\textbf{Assumption: Global Density Cap}]\label[assume]{assumption:density}
	Let $N$ be the total number of open, flow, and passage regions in $\freespace$ for some environment \environment{}. Let the maximal sphere packing\footnote{Sphere packing is difficult and well-studied \cite{scott1969density}. Similar sphere-modeled multi-agent robotics work assumes conservative estimates of density, often in the range $50-60\%$
	\cite{chinta_coordinating_motion_2018} or lower. Identifying $M$, the upper bound, in practice is non-trivial, but determining a number below $M$ that generously upholds the density assumption while outperforming the density of $50\%$ is achievable.} of a space be $M$ Then the global cap on the number of robots allowed in \environment{} is $M$ - $N$. In other words, there is always $N$ spots' worth of unoccupied space.
\end{tcolorbox}

In open space, robots can move freely; in flow space, there is an associated directional constraint on robot motion; passage regions help robots move between flow regions. 

We define the following: given a bounded environment \environment{} with free space \freespace{} and robots of radius \radius{}, we partition \freespace{} into sets \openSpace{} (the open space), \flowSpace{} (the flow space), and \passageSpace{} (the passage regions) such that:

\begin{enumerate}
	\item \flowSpace{} contains a minimally one-robot-wide corridor around every obstacle in \environment{} and the interior of \environment's perimeter.
	\item Every point in \freespace{} that can be reached by a robot without following any rules can be reached by a robot following our proposed motion rules in \cref{sec:method.motionrules} in \openSpace{}, \flowSpace{}, or \passageSpace{}.
	\item Every point in \flowSpace{} constrains robot motion to move in a particular direction (clockwise or counter-clockwise) around the perimeter and each obstacle in \environment{}, and that these directions are \emph{compatible} with each other --- that is, that there are no two adjacent points directing a robot to move in opposing directions.
	\item \passageSpace{} is composed of passage regions \passageRegion{}, where each passage region is a spot (a robot-shaped disc) that connects adjacent flow regions.
\end{enumerate}
%We present a complete algorithm for partitioning \freespace{} into \openSpace{}, \flowSpace{}, and \passageSpace{} with the above properties in \cref{appendix:B}.
%
\section{Rules of Motion}\label{sec:method.motionrules}
At any time, a robot is exclusive either in an open, flow, or passage region, or is transitioning between two regions.
These conditions are disjoint --- a robot may not be in two types of region at the same time without transitioning.
To transition between regions, a robot must use the ``space request'' mechanism defined in~\cref{sec:god}.
In the following, let $\robot \in \robots$ be a robot, $\robotState$ its position at time $t$, and $\velState$ its velocity vector at time $t$.
We identify \robot{} with its geometry, i.e.\ saying ``all of \robot{} is contained by a region'' means that the \radius-radius disc representing \robot{} is contained by the region.
\subsection{Determining a robot's region and transition state}
We determine whether \robot{} is transitioning between regions or in a particular region type at time $t$ by the following sequence of checks:
\begin{enumerate}
    \item If $\robotState \cap P \ne \emptyset$ for a passage region $P \in \passageSpace$, \robot{} is in a passage region.
    \item Else, if \robotState{} is \emph{completely} contained in a flow region $F \subset \flowSpace$, \robot{} is in a flow region.
    \item Else, if \robotState{} is \emph{completely} contained by an open region $O \subset \openSpace$ (i.e.\ $\ball_\radius(\robotState) \subseteq O$), \robot{} is in an open region.
    \item Otherwise, if none of the other conditions hold, \robot{} is transitioning between regions.
\end{enumerate}

\subsection{Constraints on robot motion}
If \robotState{} is in a flow region, its velocity \velState{} must satisfy $\velState \cdot \vec{d}_{c_i} \ge 0$, for $c_i$ the centroid of \robotState{} and $\vec{d}_{c_i}$ the flow direction at $c_i$.
In any region type, the pushing rules defined in~\cref{sec:god} apply and may constrain the motion of \robot.
If \robot{} is transitioning between regions, it must abide by all of the motion constraints for the region types it is transitioning between (e.g.\ a robot moving into a flow region cannot move against the direction of flow at its transition point).

\begin{figure}\label{fig:regions2}
	\centering
	\includegraphics[width = .8\linewidth]{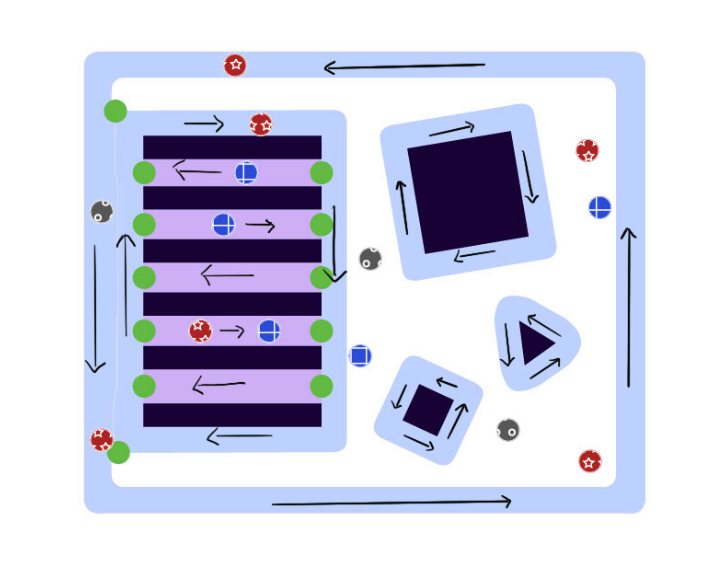}
	\caption{An example of an environment partitioned into open regions (white), flow regions (periwinkle and lilac), and passage regions (green circles) with arrows indicating direction of motion. }
	\label{fig:flowdirections}
\end{figure}

\subsection{Space Requests}\label{sec:method.spacerequests}

Space requests are the core of our adaptive centralization mechanism. 
Robots use space requests to
\begin{enumerate*}[label=(\arabic*)]
    \item transition between regions and
    \item resolve spatial conflicts (i.e.\ conflicting attempts to occupy a space)
\end{enumerate*}.
Space requests are tuples $(S, d, T)$ comprising:
\begin{description}
	\item[The primary spot $S$:] The primary spot represents the goal space, if any, that the robot wishes to access. $S$ is a radius \radius{} ball\footnote{\label{footnote:capsule}Or, in the case of a goal in or transition through a passage region, the convex hull of two adjacent radius \radius{} balls (\cref{note:exception}).} lying entirely in one region if the robot has a goal, or \texttt{null} otherwise.
	\item[The duration $d$:] A real number representing the amount of time the robot would like to spend in the primary spot.
	\item[The transition spot $T$:] The transition spot represents the space, if any, that the robot may need to access to move between two regions. $T$ is a radius \radius{} ball\cref{footnote:capsule} in a region adjacent to the robot's current location if the robot is moving between regions, or \texttt{null} otherwise.
\end{description}

\noindent{}To use space requests, each robot \robot{} maintains the following:
\begin{description}
	\item[Priority:] A tuple $\priority_i = (\timestamp_i, \prioritybreaker_i)$ where $\timestamp_i$ is a timestamp and $\prioritybreaker_i \in \mathbb{N}$ is a unique label assigned to \robot{}.
	``Timestamp priority'' refers to $\timestamp_i$ alone.
	If \robot{} does not have an active space request at some time $t$, then $\timestamp_i = \infty$.
	\item[Personal space request:] $SR_{p_i} = (S_{p_i}, d_{p_i}, T_{p_i})$, a space request used for the robot's unforced goals and transitions, as in~\cref{sec:method.srs.personal}. 
	\item[Forced space request:] $SR_{f_i} = (S_{f_i}, d_{f_i}, T_{f_i})$, a space request used for the robot's forced goals and transitions, as in~\cref{sec:method.srs.pushing}.
	$d_f$ is always zero.
	\item[Pusher identity:] $I_p$, a record of the robot (if any) which originally initiated the pushing ``network'' containing \robot{}.
	If \robot{} is not being pushed, $I_p = \prioritybreaker_i$.
\end{description}

\subsection{Moving with personal space requests}\label{sec:method.srs.personal}
Space requests for voluntary movement are a way of reserving access to a desired space.
A robot \robot{} communicates its intent to move into a space by submitting a space request, the request is checked to ensure that it is valid and won't cause a collision, and then, once granted access to the requested space, \robot{} can safely move in.
There are three stages to using a personal space request $(S_{p_i}, d_{p_i}, T_{p_i})$:
\begin{enumerate*}[label=(\arabic*)]
    \item making the request,
    \item moving into the request once it is granted, and
    \item completing the request
\end{enumerate*}.

\noindent\textbf{Making a space request:}
A robot \robot{} can make a new space request if $S_{p_i}$, the primary spot of its personal space request, is currently \texttt{null}.
To do so, \robot{} sets $S_{p_i}$ to the spot it wishes to move into, sets $d_{p_i}$ to the duration it wishes to spend in $S_{p_i}$, and communicates $SR_{p_i}$ to the space request arbiter \god{} (\cref{sec:god}).
\robot{} then waits for the arbiter to notify it that its request has been granted and set its timestamp priority to the time of approval.
While it waits, \robot{} cannot move unless forced (per~\cref{sec:method.srs.pushing}).
\robot{} can cancel its personal space request at any time by setting $S_{p_i}$ to \texttt{null} and communicating $SR_{p_i}$ to \god.

\noindent\textbf{Moving into a granted space request:}
Once $SR_{p_i}$ is granted, \robot{} must move into $S_{p_i}$ as quickly as possible, following the shortest collision-free path to the requested space (abiding by motion rules).
If this path crosses between regions, \robot{} must use the transition spot $T_{p_i}$ from $SR_{p_i}$, by setting $T_{p_i}$ to a spot 
\begin{enumerate*}[label=(\arabic*)]
    \item tangent to \robot{} and
    \item completely in the region \robot{} wants to enter
\end{enumerate*}, and communicating the modified $SR_{p_i}$ to \god{} for approval of the transition per the above.
%The transition spot requires the robot to move perpendicular to the boundary separating the robot's current region and desired adjacent region. 
Once \robot{} changes regions, the transition spot is reset to \texttt{null}.

\noindent\textbf{Completing a space request:} Once \robot{} has moved into $S_{p_i}$ (i.e. \robot{} and $S_{p_i}$ align) for $d_{p_i}$ seconds, the space request is completed and $SR_{p_i}$ is reset to $(\texttt{null}, 0.0, \texttt{null})$.

\begin{figure}\label{fig:spacerequest}
	\centering
	\includegraphics[width = .3\linewidth]{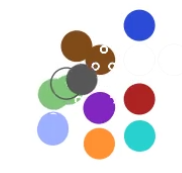}
	\caption{An example of multiple robots (green, gray, and brown) satisfying their space requests (green with polka dots, empty gray circle, and brown with polka dots) at the same time.}
	\label{fig:cluster}
\end{figure}

\subsection{Forced space requests and pushing}\label{sec:method.srs.pushing}
To guarantee progress for all robots, we sometimes need to make one or more robots move out of the space another robot wants to occupy.
We do so by introducing a notion of ``pushing'': a mechanism by which a high priority robot can clear its path.
At a high level, pushing is triggered when a higher priority robot encounters a lower priority robot in its way.
If that lower priority robot can move out of the higher priority robot's path, it must; otherwise, if its own path is blocked, it transitively pushes out, borrowing the higher priority to do so.
This iterative pushing, combined with a total ordering of robot priorities, ensures that the original higher priority robot will be able to proceed toward its goal in finite time (proving this is the primary focus of~\cref{sec:analysis}).

In more detail: a robot \robot{} can (in different circumstances) both \emph{push} and \emph{be pushed}.
Each of these processes uses each robot's priority $\priority_i$ and forced space request $SR_{f_i}$.

\noindent\textbf{Comparing priorities:}
We impose a total order on robot priorities by saying that, for $i \ne j$, $\priority_i < \priority_j$ iff $\timestamp_i$ is later than $\timestamp_j$ or $\timestamp_i = \timestamp_j$ and $\prioritybreaker_i < \prioritybreaker_j$.
This order ensures that earlier requests have higher priority, and thus that a space request's priority monotonically increases until it is completed.

\noindent\textbf{Pushing:}
When a robot \robot{} is moving to complete its space request (personal or forced) and comes tangent to\footnote{\label{footnote:tangent}Note that this does not imply physical contact between robots; \radius{} can be chosen to provide a buffer of space around each robot.} a robot $r_j$ with lower timestamp priority, then $r_i$ pushes $r_j$.
$r_j$ first sets $I_{p_j} = I_{p_i}$ and adopts the pusher's (\robot{}) timestamp priority for its forced space request $SR_{f_j}$.
We then define a ``pushing vector'' recursively to derive how $r_j$ has to move from $\robot{}$'s pushing vector. 
If \robot{} is not being pushed (i.e. it's the pushee), then its pushing vector is the vector from \robotState{} to $r_j(t)$.
If \robot{} is in a flow region, then its pushing vector is the flow direction.
Otherwise, $\robot{}$'s pushing vector also abides by the following recursive definition.
\hypertarget{pushrules}{When \robot{} pushes robot $r_j$, $r_j$ selects a unit vector $p_\theta$ with an angle ($\theta$) within $\pi/2$ of $\robot{}$'s pushing vector ($p_i$) and averages $p_\theta$ and $p_i$ to generate $p_j$.}
%Then, $r_j$ must pick a direction of movement within $\pm\frac{\pi}{2}$ radians of \robot's pushing vector abiding by motion rules. $r_j$'s new pushing vector is now the average between $\robot{}$'s pushing vector, and the vector defined between $r_j$'s initial position to its desired spot.

\noindent\textbf{Being pushed:}
A robot $r_j$ is pushed when it comes tangent to\cref{footnote:tangent} a robot \robot{} with equal or higher priority.
There is one notable exception: to ensure that no robot can transitively push itself, if the timestamp priorities of $r_j$ and $r_i$ are equal, but $I_{p_i} = \prioritybreaker_j$ (i.e.\ $r_j$ originated the pushing network containing \robot), then $r_j$ is not considered pushed.
When $r_j$ \textit{is} pushed, it sets $T_{p_j} = \texttt{null}$ (cancelling any pending requests to transition between regions for voluntary movement), and adopts \robot's timestamp priority.
$r_j$ must immediately attempt to move away from \robot{} into a spot adjacent to its current location within $\pm\frac{\pi}{2}$ radians of \robot's pushing vector. 
If there exists an unoccupied adjacent spot within the allowed angle range, $r_j$ must move into it.
If $r_j$ needs to place a space request to move (i.e.\ there are no unoccupied adjacent spots within the allowed angle range), then it makes and completes a \emph{forced space request} using $SR_{f_j}$ in the angle range, following the same process as for personal space requests with \robot's timestamp priority for the request.
A forced space request is immediately canceled if the robot is no longer being pushed or transitioning to satisfy the pushed request. $r_j$ updates its pushing vector as described \hyperlink{pushrules}{above}.

When equal priority robots push each other and neither originated the pushing network, their pushing vectors are averaged, ensuring a consistent overall pushing direction.
Pushing is re-evaluated every timestep, so pre-established redundant forced space requests are automatically canceled.
%Observe that due to the above rules, if a robot in a packed flow region is pushed, it will attempt to push \emph{out} of the flow region.

\begin{figure}
	\centering
	\includegraphics[width = .5\linewidth]{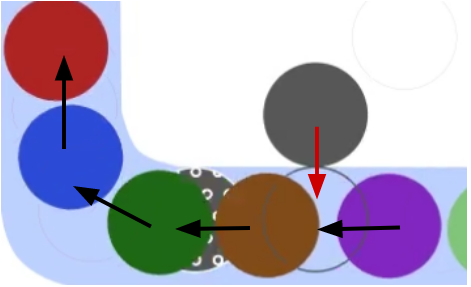}
	\caption{An example of how pushing occurs when TR (gray) is moving to granted space request (gray circle), the desired motion is shown as a red arrow. Each pushed robot's direction is indicated by an arrow (in particular, observe that the purple robot overlaps the requested spot, so it must move forward in the flow to get out of the way). }
	\label{fig:pushing}
\end{figure}

Pushing is covered in more detail in~\cref{appendix:A}.

\refstepcounter{note}
\begin{tcolorbox}[title=\textbf{The Passage Region Exception}]\label[note]{note:exception}
Space requests work differently when a robot \robot{} wants to occupy a passage region \passageRegion.
\begin{enumerate}
    \item The primary or transition spot (depending on why \robot{} wants to occupy \passageRegion) becomes a \textit{capsule} region defined by $\passageRegion\bigcup B_{EXIT}$, where $B_{EXIT}$ is a radius \radius{} ball adjacent to \passageRegion{} in the flow region \robot{} wishes to exit into.
    \item A robot pushed by \robot{} can pass through the \textit{capsule} to fulfill pushing, though it must clear the \textit{capsule} as quickly as possible.
\end{enumerate}
This allows robots to ``escape'' packed flow regions to allow \robot{} in and prevents deadlock caused by \robot{} blocking other robots from moving out of its way (see \cref{fig:figure8}).
\end{tcolorbox}

\subsection{Global Space Request Arbiter}\label{sec:god}
The global space request arbiter \god{} is in charge of space requests.
We refer to \god{} as an adaptive centralization mechanism because it coordinates precisely the set of robots which must be moved for a given robot to make progress and determines this set online.
In practice, the arbiter could be implemented as a centralized entity or through local communication directly between robots.
%The space request protocol requires us to define a notion of \emph{request priority} as well as the structure of a space request itself.
The arbiter has an initial role of randomly assigning a unique ID $\prioritybreaker_i$ to each robot \robot.
It may reassign $\prioritybreaker_i$ for all robots at fixed intervals to provide better overall fairness in access to resources (as $\prioritybreaker_i$ breaks ties when comparing priorities).
However, the arbiter's main job is to keep track of space requests and their priorities.
Every time a robot makes a space request, it is submitted to the global arbiter with the timestamp \timestamp{} when it was placed.
The arbiter keeps track of the priorities $(\timestamp_i, \prioritybreaker_i)$ of all robots that have made a space request, removes cancelled requests, and uses this information to send pushing information to the robots and grant requests. 

\god{} must approve all primary and transition spots for all space requests.
The approval process first checks that the requested spot is valid (i.e.\ that it is contained in the free space).
The arbiter immediately grants space requests that have valid spots and do not require transitioning into a flow or passage region.
Requests to transition into a flow or passage region are approved once the requested spot is clear (i.e.\ does not contain any robots).
When a request is made to transition into a flow or passage region, the arbiter informs nearby robots; robots with lower timestamp priorities must stay clear of the spot or exit it if they are currently occupying it.
If two conflicting requests (i.e.\ the requested spots overlap) with the same timestamp priority are ready to be granted, \god{} grants the request with the higher-priority value of \prioritybreaker.

\section{Analysis}\label{sec:analysis}
We will now show that following the rules defined in~\cref{sec:method.motionrules} in an environment partitioned according to~\cref{sec:method.spacetypes} suffices to guarantee progress for every robot in the shared space.

\begin{theorem}[Progress]\label{theorem:progress}
  For any valid environment \environment{} with robots \robots{}, pick (without loss of generality) some \(\robot \in \robots\) with a valid space request \spaceRequest{}.
  Then \robot{} will reach satisfy \spaceRequest{} within finite time.
\end{theorem}

We build our proof of \cref{theorem:progress} by first showing that our motion rules result in progress in a series of simplified environment types, then compose these results together to guarantee progress for all valid environments.
For space, we provide sketches of the full proofs.

\begin{lemma}\label{lemma:priority.resolution}
Assume any space request is satisfied in finite time.
Then every space request will eventually have the highest priority.
\end{lemma}

\begin{sketch}
    Each robot can have at most one personal space request and one forced space request a time.
    The global space request arbiter processes space requests in order of submission.
    Thus, there are a finite number of requests, processed in order of arrival, and (by assumption) each is completed in finite time.
    Further, by the ordering defined on request priorities, a request's priority
    \begin{enumerate*}[label=(\arabic*)]
        \item is monotonically increasing and
        \item increases whenever a previously submitted request is completed
    \end{enumerate*}.
    Therefore, every space request will eventually have the highest priority.
\end{sketch}

We introduce the following symbols for proof sketch brevity.
Let $\gamma$ be a finite set of transitively adjacent open, passage, and flow regions in an environment \environment{}, and let $\Gamma \subseteq \environment$ be the set of all possible $\gamma$ given environment \environment.
Finally, let TR signify the robot \robot{} with current global highest priority.

\begin{lemma}\label{lemma:infinite.open}
Let $\Gamma'$ be  a single infinite obstacle-free open region, and note that $|\robots|$ is finite.
Let the current TR be robot \robot{}, and let \robot{} be a finite distance away from its granted space request \spaceRequest{}.
Both \robot{} and \spaceRequest{} lie in $\Gamma'$.
Then \robot{} will complete \spaceRequest{} in finite time.
\end{lemma}

\begin{sketch}
Assume $\Gamma'$ and \robot{} as given.
We will show that \robot{} will satisfy its request in finite time by induction on $|\robots|$.

\noindent\textbf{Base case:} 
Let $|\robots| = 2$, containing TR \robot{} and one other robot $r_j$.
There can be at most one robot blocking the straight line path from \robotState{} to its goal.
By the rules for being pushed, and since \robot{} is TR, the obstructing robot $r_j$ will move away from \robot{} immediately if there is an adjacent open spot.
By construction of $\Gamma'$, there must be an empty spot adjacent to $r_j$.
Therefore, $r_j$ will always clear \robot's path in finite time and thus \robot{} will always complete \spaceRequest{} in finite time.

\noindent\textbf{Inductive Hypothesis:} Let $|\robots| \leq k$ for $k \in \mathbb{N}$.
Then TR \robot{} will complete \spaceRequest{} in finite time. 

\noindent\textbf{Inductive Step:} Let $|\robots| = k + 1$.
There are at most $k$ robots that lie within the straight line path from \robot{} to \spaceRequest{}.
Robots on this path may have their movement blocked by at most $k-1$ robots.
Since $\Gamma'$ is infinite, there always exists an empty spot transitively adjacent to every $r \in \robots$.
By the pushing rules, the blocking robots directly adjacent to said empty spot must immediately move into it.
Thus, combining this motion with the inductive hypothesis, each robot obstructing \robot's path is able to and will in finite time move out of the way.
 
\noindent\textbf{NOTE:} As described in \hyperlink{pushrules}{the pushing rules} the pushing vector of the $n$th robot in a pushing chain is defined as $p_n = \frac{p_{n-1}+p_\theta}{2}$, where $p_{n-1}$ is the pushing vector of the robot that pushed $p_n$.
This sequence of vectors' angles in the global frame is bounded above by the original top robot's pushing vector angle $\pm \pi/2$.
This ensures that no chain of pushed robots will ever circle back towards the original pusher (TR) and close a cycle with the original TR (which would preclude progress).
This process is also described in \cref{appendix:A}.
\end{sketch}

\begin{lemma}\label{lemma:finite.time.transition.region.request}
     Let $\Gamma'$ be the subset of $\Gamma$ such that each $\gamma \in \Gamma'$ consists of only a single flow network --- a maximal subset of $\flowSpace{}\bigcup \passageSpace{}$ for which all regions are adjacent to at least one other region in the subset. Let \robot{} be TR with space request \spaceRequest{} through a passage region \passageRegion{} to exit into a flow region \flowRegion{}.
    \robot{} will complete \spaceRequest{} in finite time.
\end{lemma}

\begin{figure}
	\centering
	\includegraphics[width = .7\linewidth]{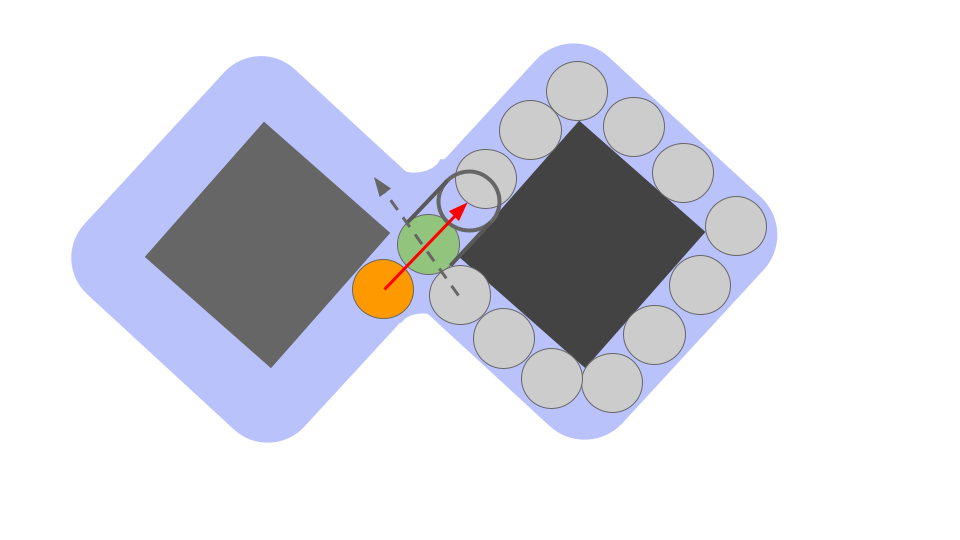}
	\caption{An example of a figure-eight environment where the top robot (orange) needs to clear the capsule space (outlined in gray) that overlaps the passage region (green).
	The flow region the capsule space overlaps is full of robots, so the last robot must pass through the \textit{capsule} to enter the adjacent flow region (dashed arrow).}
	\label{fig:figure8}
	\vspace*{-.5 cm}
\end{figure}

\begin{sketch}
Note that (per the \textit{passage region exception} in~\cref{note:exception}) space requests in passage regions reserve a \textit{capsule} region $C$ instead of a radius \radius{} ball. The capsule overlaps both the passage region \passageRegion{} and an open spot in the exit flow region adjacent to the capsule. We will call this open spot $B_{exit}$ and the flow region it belongs to $\flowRegion{}_{exit}$. $\flowRegion{}_{exit}$ flows in one direction by construction, moving away from \passageRegion{}.
The proof proceeds by cases:

\noindent\textbf{Case 1:} C is already clear.
\robot{} can move into its requested passage region and exit region immediately.

\noindent\textbf{Case 2:} There are a finite set of robots ordered in proximity to \passageRegion{} $\{r_1, r_2,..., r_j, ..., r_k\}$ in such that $\{r_1,...,r_j\}$ lie in $B_{exit}\bigcup C$ where $i\leq k$. A capsule can intersect at most 3 robots at once, so there must be at least $j$ robots worth of unoccupied space in $\gamma \in \Gamma'$ because of \cref{assumption:density}.
\\
\hypertarget{thesentence}{If the flow region has $j$ unoccupied robot's worth of space, robots $\{r_{j+1}, ..., r_k\}$ moving forward in the flow will cause the $j$ unoccupied robot's worth of space to move behind $r_{i+1}$ and thus provide enough room for $\{r_1,...,r_j\}$ to empty $C$.} 
If the flow region does not have $j$ unoccupied robot's worth of space, then $j$ robots need to exit $\flowRegion{}_{exit}$ to another flow region through a passage region. 
If the adjacent flow regions have j unoccupied robot's worth of space, each robot leaving $\flowRegion{}_{exit}$ will use TR's priority to reserve the passage region and transition into the adjacent flow regions and like \hyperlink{thesentence}{the prior argument} there will be enough room made for the $j$ robots to clear $C$. 
In the worst case, pushing loops back to \robot{}, meaning that a robot $\not \in \{r_1,...,r_j\}$ must pass through C to allow $\{r_1,...,r_j\}$ to clear the capsule-adjacent spot.
By the passage region exception (\cref{note:exception}), this can and will happen.
This is important for scenarios like that shown in \cref{fig:figure8}.
\end{sketch}

\begin{lemma}\label{lemma:single.flow}
    Let $\Gamma'$ be the same as in \cref{lemma:finite.time.transition.region.request}.
    Let the current TR be \robot{} in $\Gamma'$ with space request $z\in \Gamma'$.
    \robot{} will reach \spaceRequest{} in finite time without exiting $\Gamma$.
\end{lemma}

\begin{sketch}
A flow network can be represented as a network graph where nodes are unique flow regions and directed edges are transition directions across boundaries through passage regions.
By construction, all flow regions are connected transitively to all other flow regions.
Thus, flow networks are strongly connected graphs.
\cref{lemma:finite.time.transition.region.request} establishes that passage regions can always be used in finite time, so robots can move between any two flow regions in finite time.
By the flow motion rules, robots also move through each flow region in finite time.
Therefore, \robot{} can reach \spaceRequest{} without leaving $\Gamma'$, in finite time.
\end{sketch}

\begin{lemma}\label{lemma:one.flow.open}
Let $\Gamma'$ be the subset of $\Gamma$ such that each $\gamma \in \Gamma'$ is a flow region network and a single open region adjacent to the flow region network.
Then TR \robot{} will reach its granted space request $\spaceRequest \in \Gamma'$ in finite time, without exiting $\Gamma'$.
\end{lemma}

\begin{sketch}
%This proof uses~\cref{note:finite.time} to show several cases:
%\\
\refstepcounter{sublemma}
\textbf{Sub-Lemma \thesublemma: Region transitions take finite time}\label[sublemma]{note:finite.time}
\textit{Claim: If $r_j$ is TR or is being pushed by TR \robot{}, $r_j$ can always transition regions in finite time.} 
There are three cases to consider, by transition type:

\textit{(Flow $\rightarrow$ Flow):}
\cref{lemma:finite.time.transition.region.request} makes the claim hold for $r_j$ moving between flow regions via a passage region. 

\textit{(Flow $\rightarrow$ Open):}
If there is not enough room for $r_j \in \flowSpace$ to move into an adjacent spot in \openSpace{}, it will push outward into \openSpace.
If there is sufficient unoccupied space in \openSpace, then this pushing will cause robots to occupy as much of this space as possible -- gathering together into a more densely packed group-- to try to clear a spot for $r_j$ in finite time.
Otherwise, the pushing network will propagate until a robot is pushed into \flowSpace, clearing a spot for $r_j$ in finite time.
These cases are exhaustive by \cref{assumption:density}.

\textit{(Open $\rightarrow$ Flow):}
If the flow network is not full, \cref{lemma:single.flow} means that $r_j$ can use a space request to transition in finite time.
If the network \textit{is} full, \cref{assumption:density} implies that there is a free spot in the open region.
Thus a robot can exit the full flow network to make room for $r_j$ to enter; by the rules for pushing (which causes robots in the flow network to push out into open space), this will happen in finite time.

\noindent\textbf{Case 1:}
\spaceRequest{} is in the open region \openRegion{}.

\noindent\textit{(\robot{} in \openRegion{}):}
We use the same inductive structure as \cref{lemma:infinite.open} with a different base case.
An unoccupied spot somewhere in $\gamma$ is guaranteed by \cref{assumption:density}. Either this unoccupied spot is in \openRegion{} and pushing ensures the space request will be completed in finite time (by condensing the robots in \openRegion), or the spot is in the adjacent flow network and pushing ensures an obstructing robot will be pushed into the flow network allowing the space request to be completed in finite time.

\noindent\textit{(\robot{} in \flowRegion{}):}
If \robot{} is in the flow network and its space request is in \openRegion, \robot{} can reach a spot adjacent to \openRegion{} by \cref{lemma:single.flow}.
Then, by \cref{note:finite.time}, it can enter \openRegion{} in finite time.

\noindent\textbf{Case 2:}
\spaceRequest{} is in a flow region \flowRegion{}.

\noindent\textit{(\robot{} in \flowRegion{}):}
By~\cref{lemma:single.flow}, \robot{}  can reach its request.

\noindent\textit{(\robot{} in \openRegion{}):}
\robot{} can make it to a spot adjacent to the flow network via Case 1, and then transition across the boundary by \cref{note:finite.time}.
\robot{} can then follow the flow network motion as per \cref{lemma:single.flow} to make it to its space request.
\end{sketch}

\begin{corollary}\label{corollary:finite.flow.open}
Let $\Gamma'$ be the subset of $\Gamma$ such that each $\gamma \in \Gamma'$ is a finite number of transitively adjacent flow region networks and transitively adjacent open regions.
The TR \robot{} will reach its requested spot in finite time.
\end{corollary}
\begin{sketch}

\refstepcounter{observation}
\noindent\textbf{Observation \theobservation: Full Flow Region}\label[observation]{note:full.flow}
In a packed flow network, the pushing rules imply that all robots will join \robot's pushing network until one can exit into an adjacent region.

\refstepcounter{observation}
\noindent \textbf{Observation \theobservation: Full Open Region}\label[observation]{note:full.open}
In a packed open region, a robot originating a pushing network will push all robots on one side of the plane orthogonal to the robot's pushing vector until a robot is able to exit into an adjacent flow region.
At this stage, it is not known if the whole open region is packed.
If the pushing network pushes back into the same open region, all robots in the region will be included in the pushing network and the region is known to be full.
This follows because the initial pushing vector and the pushing vectors re-entering the open region will have an angle of less than or equal to $\frac{\pi}{2}$ radians between them, creating ``pushing planes'' that cover the open region.

\cref{lemma:one.flow.open} ensures that if there is an unoccupied spot in a region adjacent to the TR, pushing can and will use this spot for \robot{} to achieve its goal.
\cref{assumption:density} ensures there are $N$ unoccupied spots in the space but does not guarantee they are --- at any given time --- in a region adjacent to the TR.
However, in a $\gamma$ with more than one flow network adjacent to an open space, we need to prove that as pushing propagates, the unoccupied space guaranteed by \cref{assumption:density} will not move in cycles between regions without ever being moved to the region where \spaceRequest{} lies.
By construction of $\gamma$, all regions are transitively adjacent to all other regions in $\gamma$.
A pushing network will push into any adjacent unoccupied spots, or expand the pushing network to encompass any adjacent robots.
Therefore, \cref{note:full.flow} and \cref{note:full.open} mean that pushing will always grow to cover larger subsets of the environment and find any possible unoccupied spots.
 \end{sketch}

 \begin{sketch} \textbf{Proving Theorem 1 (Progress)}
From \cref{corollary:finite.flow.open} we know that any TR will reach its space request in finite time.
From \cref{lemma:priority.resolution} we know if all space requests are satisfied in finite time, all robots with a space request become TR in finite time.
Thus, space requests will be satisfied in finite time, and progress for all robots is guaranteed.
\end{sketch}   
%\wtnote[inline]{Also want to say something informally about communication/computation overhead.}

\section{Evaluation}\label{sec:evaluation}
We have implemented a proof-of-concept simulation of our proposed approach, as shown in~\cref{fig:pushing,fig:cluster,fig:flowdirections}.
In our supplementary material, we include a video demonstrating how robots move, resolve contention for space, and push in packed environments when abiding by our motion rules.
Each robot in our simulation creates and follows their own plan without knowledge of other robots' planning details.
We show examples of several possible polygonal environments partitioned as per~\cref{sec:method.spacetypes}.

\section{Conclusion}\label{sec:conclusion}
In this work we present a method for ensuring progress for every robot in a shared multi-robot space via a \emph{motion pact}.
Our motion pact is comprised of a partitioning of the environment freespace along with a set of motion rules that allow robots to plan independently and flexibly, with protection from being blocked (either intentionally or inadvertently) by other robots' behavior.
We prove that the rules guarantee progress for every robot, and demonstrate a proof-of-concept implementation of our system.

This work assumes that all robots are holonomic and capable of instantaneous bounded acceleration. 
Beyond relaxing this assumption, this work leaves several exciting open questions.
For example, one could consider re-imagining the form of goals to encompass team-based tasks, where multiple robots must collaborate to satisfy their objectives rather than individually reach a location. With such an expansion, robots can work in teams, or even switch between teams, to more flexibly move and work in a shared space.

\appendices
\renewcommand{\thesectiondis}[2]{\Alph{section}:}
\section{Pushing Logic}\label[app]{appendix:A}
We now describe, in \cref{algo:bkwd}, the logic for computing the pushing angles and velocity vectors for each robot in the set \robots{} at a timestep $t$.
\Cref{algo:bkwd} depends on the following subroutines:

\begin{itemize}
	\item \texttt{activeSR(\robot)}: Returns the first non-null space request for \robot, in the order $T_f$, $S_f$, $ T_p$, $S_p$.
	\item \texttt{adoptpriority($r_j$, \robot{})}: Copies the timestamp $\tau$ from \robot's priority and the identity $r_p$ of the original pushing robot for \robot{} to $r_j$.
	\item \texttt{arc($push_i$)}: Returns the set of angles within $\pm\frac{\pi}{2}$ radians of $push_i$.
	\item \texttt{overlaps($d_1, d_2$)}: Returns \texttt{true} iff the discs $d_1$ and $d_2$ intersect. If called with robots $r_i, r_j$ as arguments, $d_1$ and $d_2$ are the \spot-sized discs centered at the positions of $r_i$ and $r_j$, respectively.
\end{itemize}

%The pushing logic, then, is as follows:
\begin{algorithm}[h]
	\SetAlgoLined
	\SetKwInOut{Input}{Input}
	\SetKwInOut{Output}{Output}
	\SetKwProg{Initialization}{Initialization}{}{}
	\SetKw{Continue}{continue}
	\SetKw{And}{and}
	\DontPrintSemicolon

	\While{any $r_i\in\robots$'s velocity changes}{
		\For{$\robot, r_j\in \robots{} \times \robots{}$}{
			\lIf{$\priority_i < \priority_j$ or $I_{p_i} = \prioritybreaker_j$}{
				\Continue{}
			}
			\ElseIf{$priority(r_i) = priority(r_j)$ \And $(r_j(t) - \robotState) \in arc(push_j)$}{
				$push_j = \frac{push_i + push_j}{2}$\;
				\Continue{}
			}

			\lIf{$overlaps(\robotState, r_j(t))$}{ $p = \robotState$ }

			\If{overlaps(activeSR(\robot{}),\robot{}) \And overlaps(activeSR(\robot{}), $r_j(t)$) }{
				$p = centroid(activeSR(\robot{}))\;$
			}

			$\Theta = cos^{-1}(\frac{push_i \cdot (r_j(t)-p)}{\|push_i\| \|r_j(t)-p \|})$\label{line:theta}\;
			\If{$\Theta \in arc(push_i)$}{
				$adoptpriority(r_j, \robot{})$\;
				\lIf{$r_j \in \flowSpace$}{$v_j = d_{r_j(t)}$}
				\lElse{$v_j= r_j(t)-\robotState$}
			}
		}
	}
	\caption{Computing pushing vectors}
	\label{algo:bkwd}
\end{algorithm}

At a high level, \cref{algo:bkwd} iterates through each pair of robots $r_i, r_j \in \robots$ checking for each $r_j$ that is close enough to a higher-priority $r_i$, or the active space request of a higher-priority $r_i$, for pushing to engage.
We compute the angle between the velocity vector $push_i$ of $r_i$ and the vector pointing from $r_i$ (or its active space request) to $r_j$ (\cref{line:theta}), and, if this angle is within the acceptable $\pm\frac{\pi}{2}$ range of the original pushing vector, propagate the pushing priority of $r_i$ to $r_j$ and set $r_j$'s velocity vector to move out of the way.
This process repeats until no robot's velocity vector changes, at which time the robots in \robots{} move accordingly.

\section{Region Generation}\label[app]{appendix:B}

As described in \cref{sec:method.spacetypes}, our approach requires partitioning the free space \freespace{} of a bounded environment \environment{} into three sets: the open space \openSpace{}, the flow space \flowSpace{}, and the passage region space \passageSpace.
In this section, we describe constructing such a partition.
We will refer to the set of all obstacles and the internal perimeter of the environment as \emph{flow generating objects}.
Our algorithm proceeds through the following high-level steps (with further detail in corresponding subsections):
\begin{enumerate}
	\item Construct \emph{initial flow regions} around each flow generating object to capture a space wide enough for a line of robots around each object's perimeter (\cref{sec:initial.flow.regions}).
	\item Separately construct \emph{inflated flow regions} around each flow generating object (\cref{sec:initial.flow.regions}).
	      These regions capture subsets of the free space too narrow to hold three robots side by side.
	      \begin{enumerate}
		      \item In areas where two inflated flow regions overlap, split the region lengthwise between corresponding flow generating objects.
	      \end{enumerate}
	\item Construct \emph{single-lane regions} from areas where two initial flow regions overlap (\cref{sec:single.lane.regions}) to capture flow space areas too narrow for two robots side by side.
	\item Place \emph{passage regions} (\cref{sec:pasage.regions}) to connect different flow regions in \flowSpace{}.
  \item Smooth the perimeter of open space (\cref{sec:smooth}).
	\item Assign directions to each flow region (\cref{sec:direction.assignment}).
	\item Return the flow regions as \flowSpace, the passage regions as \passageSpace, and the remainder of the free space as \openSpace.%\footnote{* and \textdagger  ensure that everywhere in \openSpace{} can hold a robot}
\end{enumerate}
% WT: We could probably save some space here by using something other than subsection, e.g. manual
% \textit stuff.
\newcommand{\appsec}[1]{\refstepcounter{subsection}\noindent\textbf{\Alph{subsection}. #1}}
\appsec{Initial Flow Regions:}\label[app]{sec:initial.flow.regions}
The initial flow regions are constructed by taking the Minkowski sum of each object $O_i$ and a \spot-sized disc.
Each flow region is $f_i = (O_i \oplus \spot) \setminus O_i$.
If any $f_i$ intersects itself or a flow generating object, the environment is invalid and we return an error.

\appsec{Inflated Flow Regions:}\label[app]{sec:inflated.flow.regions}
The inflated flow regions are constructed the same as the initial flow regions, but using a $1.5 * \spot$-sized disc.
If two inflated flow regions overlap, generating initial flow regions without modification creates a subset of \openSpace{} too narrow to hold a robot.
Instead, we split the inflated region overlaps along their central axis and add each divided-up area onto their corresponding initial flow regions.

\appsec{Single-lane Regions:}\label[app]{sec:single.lane.regions}
Anywhere that two initial flow regions overlap is a subset of the flow space too narrow to hold two robots: both initial flow regions cannot exist in the same space.
We construct single-lane regions by joining the two generating objects at the ends of each overlap and creating a new flow region between these bounds.

\appsec{Passage Regions:}\label[app]{sec:pasage.regions}
We place a passage region (per~\cref{sec:method.spacetypes}) on both ends of each single-lane region and at each point where three or more flow regions meet.
This
\begin{enumerate*}[label=(\arabic*)]
	\item ensures that transitioning for single-lane regions is controlled by space requests and
	\item covers ``sharp corners'' in flow regions.
\end{enumerate*}

\appsec{Smoothing:}\label[app]{sec:smooth}
Convolve a \spot-sized disc with the inner perimeter of the open space (i.e.\ $\freespace \setminus \flowSpace$).
Anywhere that this disc is in contact with the inner perimeter at two points, we expand the corresponding flow regions out to meet the disc.
Smoothing ensures that no reachable point in \freespace{} is made unreachable by the flow/open space partition.

\appsec{Direction Assignment:}\label[app]{sec:direction.assignment}
We assign directions to each flow region by winding either clockwise or counterclockwise around the centroid of its generating object (for single-lane regions, we arbitrarily choose one of the generating objects).
% We intersect each of the (potentially expanded) initial flow regions, single-lane regions, and passage spots with \freespace.
We then return the flow, single-lane, and passage regions.

\renewcommand*{\bibfont}{\small}
\printbibliography{}
\end{document}